\documentclass{article}

\usepackage{microtype}
\usepackage{graphicx}
\usepackage{subfigure}
\usepackage{booktabs} 

\usepackage{xcolor}

\usepackage{arydshln}

\usepackage{hyperref}

\usepackage{pifont}
\newcommand{\cmark}{\ding{51}}%
\newcommand{\xmark}{\ding{55}}%

\usepackage[accepted]{icml2024}

\usepackage{amsmath}
\usepackage{amssymb}
\usepackage{mathtools}
\usepackage{amsthm}

\usepackage[capitalize,noabbrev]{cleveref}

\theoremstyle{plain}

\theoremstyle{definition}

\theoremstyle{remark}

\usepackage[textsize=tiny]{todonotes}

\icmltitlerunning{Technical Report.}

\begin{document}

\twocolumn[
\icmltitle{\textsc{Q-Align}: Teaching LMMs for Visual Scoring via Discrete Text-Defined Levels}



\icmlsetsymbol{equal}{$\spadesuit$}
\icmlsetsymbol{lead}{$\heartsuit$}
\icmlsetsymbol{corresponding}{$\diamondsuit$}

\begin{icmlauthorlist}
\icmlauthor{Haoning Wu}{equal,lead,yyy}
\icmlauthor{Zicheng Zhang}{equal,sch}
\icmlauthor{Weixia Zhang}{sch}
\icmlauthor{Chaofeng Chen}{yyy} \\
\icmlauthor{Liang Liao}{yyy}
\icmlauthor{Chunyi Li}{sch}
\icmlauthor{Yixuan Gao}{yyy,sch}
\icmlauthor{Annan Wang}{yyy}
\icmlauthor{Erli Zhang}{yyy} \\
\icmlauthor{Wenxiu Sun}{comp}
\icmlauthor{Qiong Yan}{comp}
\icmlauthor{Xiongkuo Min}{sch}
\icmlauthor{Guangtao Zhai}{corresponding,sch}
\icmlauthor{Weisi Lin}{corresponding,yyy}
\end{icmlauthorlist}

\icmlaffiliation{yyy}{Nanyang Technological University}
\icmlaffiliation{comp}{Sensetime Research}
\icmlaffiliation{sch}{Shanghai Jiao Tong University}

\icmlcorrespondingauthor{Guangtao Zhai}{zhaiguangtao@sjtu.edu.cn}
\icmlcorrespondingauthor{Weisi Lin}{wslin@ntu.edu.sg}

\icmlkeywords{Machine Learning, ICML}

\vskip .3in
]



\printAffiliationsAndNotice{$^\spadesuit$ Equal contribution.$^\heartsuit$ Project Lead.} 

\begin{abstract}
The explosion of visual content available online underscores the requirement for an accurate machine assessor to robustly evaluate scores across diverse types of visual contents. While recent studies have demonstrated the exceptional potentials of large multi-modality models (LMMs) on a wide range of related fields, in this work, we explore how to teach them for visual rating aligned with human opinions. Observing that human raters only learn and judge \textbf{discrete text-defined levels} in subjective studies, we propose to emulate this subjective process and teach LMMs with text-defined rating levels instead of scores. The proposed \textbf{\textsc{Q-Align}} achieves state-of-the-art performance on
\textit{image quality assessment} (IQA), \textit{image aesthetic assessment} (IAA), as well as \textit{video quality assessment} (VQA) tasks under the original LMM structure. With the syllabus, we further unify the three tasks into one model, termed the \textbf{\textsc{OneAlign}}. In our experiments, we demonstrate the advantage of the discrete-level-based syllabus over direct-score-based variants for LMMs.
Our code and the pre-trained weights are released at \href{https://github.com/Q-Future/Q-Align}{\textit{https://github.com/Q-Future/Q-Align}}.
\end{abstract}

\section{Introduction}

\textit{There is always a need to score an image.} From the early focus on factors related to compression, transmission, and image processing~\cite{liveiqa}, to directly addressing user-generated content~\cite{videval} (\textit{e.g.} photos and videos taken with smartphones~\cite{spaq}), and moving on to the recently popular AI-generated content~\cite{agiqa3k}, at every stage, accurately evaluating visual content remains an indispensable need to the computer vision field. To address this need, from handcraft approaches~\cite{niqe,brisque} to deep-neural-network-based methods~\cite{nima,dbcnn,musiq}, the endeavor to improve the accuracies of visual assessors never stops. Nevertheless, while existing methods can already achieve remarkable accuracies on specific datasets by {regressing} from the mean opinion scores (MOS), the complicated factors that affect the final score in contrast with the limited capacity of these methods have resulted in their poor out-of-distribution (OOD) generalization abilities. This makes them struggle to accurately score novel types of content. Moreover, they usually experience compromised performance while handling different scoring scenarios (\textit{e.g. mixing multiple datasets}) together, making it challenging to train a unified model for different situations.

\begin{figure}
    \centering
    \includegraphics[width=\linewidth]{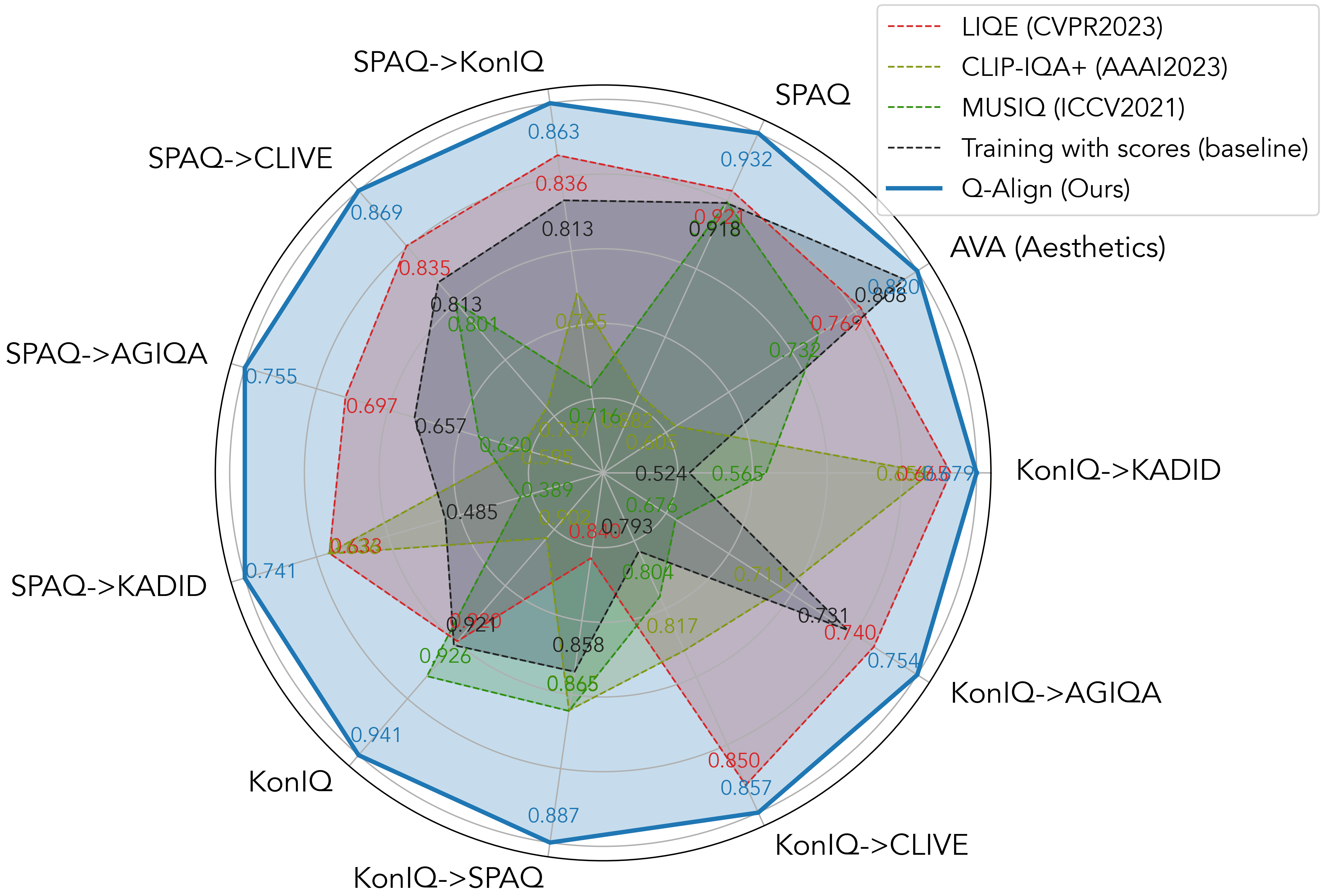}
    \vspace{-2.em}
    \caption{The \textbf{\textsc{Q-Align}} (\textit{training LMMs with text-defined levels}) in comparison with its baseline  (\textit{training LMMs with scores}) and existing state-of-the-arts, showing exceptional improvements especially on \textit{cross-set} settings. Metrics are the (SRCC+PLCC)/2. }
    \label{fig:radar}
        \vspace{-1.4em}
\end{figure}
\begin{figure*}
    \centering
    \includegraphics[width=\linewidth]{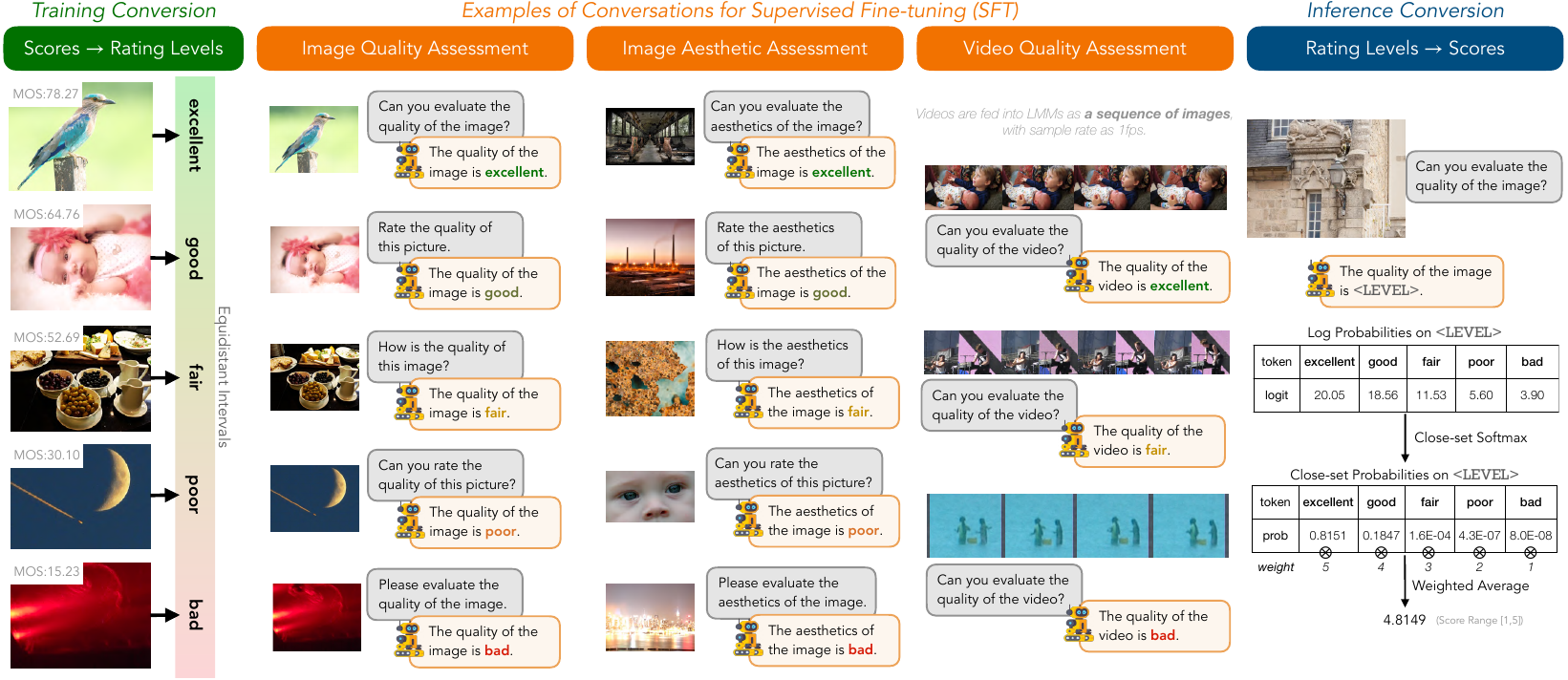}
    \vspace{-1.4em}
    \caption{The syllabus of \textbf{\textsc{Q-Align}}. Based on the general principle to teach LMMs with text-defined rating levels, we generate the instruction-response pairs by converting \textbf{existing score labels} in image quality assessment (IQA), image aesthetic assessment (IAA) and video quality assessment (VQA) datasets. During inference, by simulating the process of collecting mean opinion scores (MOS) from annotators, we extract the close-set probabilities of rating levels and perform weighted average to obtain the LMM-predicted score.}
    \vspace{-0.65em}
    \label{fig:syllabus}
\end{figure*}

In contrast, recently emerging large multi-modality models (LMMs) have shown very strong background knowledge on a wide range of visual and language disciplines. They can well understand high-level visual contents~\cite{improvedllava,mplugowl}, and effectively perceive low-level visual attributes~\cite{xcomposer}, and more importantly possess reasoning ability benefited from their strong language decoder~\cite{mmbench}. While these abilities are proved fundamental to a more accurate and robust visual scorer, existing studies~\cite{wu2023qbench} have proven that they still fall short on accurately predicting scores that are consistent with human preferences. Therefore, in our study, we investigate the important one last mile for them:

\textit{How to teach LMMs to predict scores aligned with human?}

To design the most effective syllabus, we reviewed the standard process for collecting MOS from human~\cite{itu}: First, organizers need to define several rating levels (\textit{e.g.} `\textit{excellent}', `\textit{fair}', `\textit{bad}') and select examples for each level, aligning human annotators to the standards of each level. Referring to these levels, humans mark their ratings either through a choice button or a grade-guided slider. In other words, \textbf{human annotators never learns or marks a specific score} (\textit{e.g.} \textit{3.457} in range [1,5]). Instead, these final scores are derived from the distributions of human ratings.

Meanwhile, as observed by recent explorations~\cite{wu2023qbench}, LMMs have similar \textit{behaviour patterns} to humans while instructed to score: they prefer to \textbf{respond with text-defined levels} (\textit{good/poor}); even while explicitly requested to predict numerical scores, the accuracy is significantly lower compared to deriving from levels. Therefore, it might not be optimal to directly tune LMMs to output scores.

Given the above observations, we propose a human-emulating syllabus to teach LMMs for visual scoring (the \textbf{\textsc{Q-Align}}), as shown in Fig.~\ref{fig:syllabus}. \textbf{During training}, simulating the process of training human annotators, we convert the MOS values to five text-defined rating levels~\cite{itu} (\textit{excellent/good/fair/poor/bad}), which are further formatted into instruction-response pairs, to conduct visual instruction tuning~\cite{llava} on LMMs. \textbf{During inference}, simulating the strategy to collect MOS from human ratings, we extract the log probabilities on different rating levels, employ softmax pooling to obtain the close-set probabilities of each level. Finally, we get the  LMM-predicted score from a weighted average on the close-set probabilities.

While the proposed syllabus requires only existing scores and uses even less information, it has proved far better performance than using scores as learning targets. It reaches \textbf{state-of-the-art performance on 12 datasets} of three representative visual scoring tasks with notable improvements: \textit{image quality assessment} (IQA), \textit{image aesthetic assessment} (IAA), and \textit{video quality assessment} (VQA), with especially significant improvements on unseen (OOD) datasets. 

Besides achieving state-of-the-art, the proposed \textbf{\textsc{Q-Align}} also have two exciting characteristics: \textbf{1) Data Efficiency.} It can be competitive with current state-of-the-arts with only 1/5 (IQA) or even 1/10 (IAA) data used. This could be especially useful as data collection is rather expensive for visual scoring tasks. \textbf{2) Free Combination of Datasets.} With the strong capacity of LMMs, unlike existing methods that usually face performance drop while mixing datasets~\cite{liqe}, it can freely combine different datasets for training even from different tasks (\textit{i.e. IQA and VQA}), and receive positive performance gain. With this characteristic, we propose the  \textbf{\textsc{OneAlign}}, which combines IQA, IAA and VQA datasets for training. The \textbf{\textsc{OneAlign}} is exceptionally capable on all three tasks under one unified model, with further enhanced generalization on unseen datasets.

Our core contributions can be summarized as three-fold: 
\begin{itemize}
    \item \textbf{An effective syllabus to teach LMMs to score.} Emulating from human opinion collection process, the proposed \textbf{discrete-level-based} syllabus proves its effectiveness over the \textit{score-based} variant (\textbf{+10\%}). We expect it a general strategy for training LMMs to score.
    \item \textbf{A family of more capable visual assessors.} The proposed \textbf{\textsc{Q-Align}} achieves state-of-the-art accuracy and generalization ability on multiple visual assessing tasks. It also proves competitive performance with fewer data used, and can converge with fewer training iterations.
    \item \textbf{A unified model for visual scoring.} With IQA, IAA, and VQA effectively learned independently under the same structure, we further propose \textbf{\textsc{OneAlign}}, that unifies all three tasks under one model. We hope this may open a new paradigm for visual scoring tasks.
\end{itemize}

\section{Related Works}

\paragraph{Image Quality Assessment (IQA).}
Image quality assessment (IQA) mainly focuses on the impact of distortions and other quality issues in images on human perception. Early IQA algorithms usually operate on handcraft features following the prior knowledge of statistics disciplines~\cite{ssim,brisque,niqe}. As distortion diversifies and visual content becomes more complex, data-driven end-to-end deep neural networks are increasingly applied in the IQA field, as represented by NIMA~\cite{nima}, DBCNN~\cite{dbcnn}, and HyperIQA~\cite{hyperiqa}. Following this path, MUSIQ \cite{musiq} designs a multi-scale input structure that advances the accuracy on IQA via transformers. In recent years, several methods have investigated the vision-language correspondence embedded in CLIP~\cite{CLIP} to improve generalization ability in IQA. Among them, CLIP-IQA+~\cite{clipiqa} designs a few-shot learning scheme via CoOp~\cite{coop}, and LIQE~\cite{liqe} further develops a multitask learning scheme based on CLIP. Nevertheless, they typically rely on visual-text similarity to predict quality scores, which limits their performance to be slightly inferior compared with pure visual methods. Instead, the proposed \textbf{\textsc{Q-Align}} can significantly advance state-of-the-arts on IQA, while simultaneously further improving OOD generalization ability.


\paragraph{Image Aesthetic Assessment (IAA).}
In comparison with IQA, image aesthetic assessment (IAA)~\cite{avaiaa} is a more complicated task for visual scoring. While visual quality is also considered influential to visual aesthetics, the higher-level visual attributes, such as \textit{content, lighting, color, composition}~\cite{aadb} are considered more important for IAA. As a result, deep-neural-network-based methods predominate IAA, such as NIMA and MLSP~\cite{mlspiaa}. Similar as IQA, VILA~\cite{vila} advances IAA performance by learning vision-language correspondence between images and aesthetic comments~\cite{avacomments} through a joint constrastive and captioning pretraining~\cite{coca}. Based on LMMs with rich prior knowledge, the proposed \textbf{\textsc{Q-Align}} can remarkably outperform CLIP-based approaches \textit{without} extra pre-training.

\begin{figure*}
    \centering
    \includegraphics[width=\linewidth]{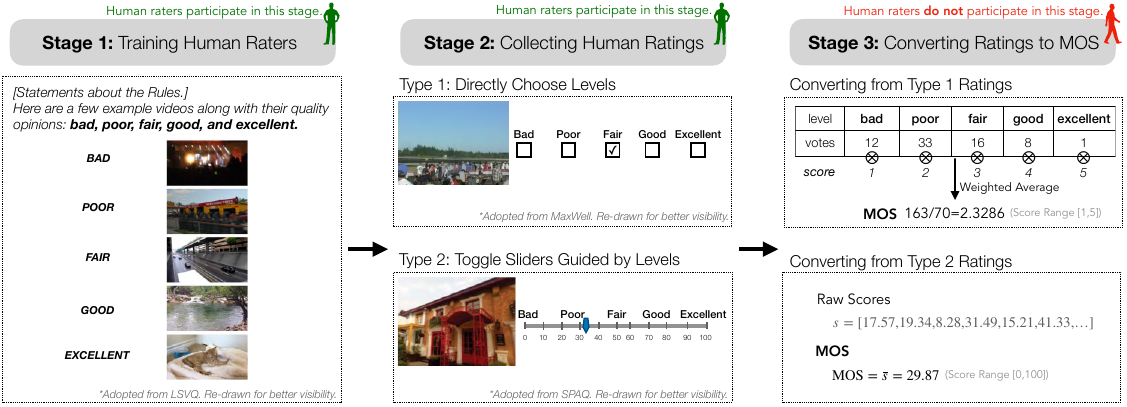}
    \vspace{-1.7em}
    \caption{[\textit{Insight 1}] \textsc{How do humans rate?} Typically, it include three stages: \textbf{(1)} Training human raters with text-defined rating levels. Simulating this, we propose the rating-level-based syllabus for LMMs. \textbf{(2)} Collecting human ratings. Human raters choose levels (Type 1) or toggle level-guided sliders to score (Type 2), \textit{without directly inputting the score in either way}. \textbf{(3)} Converting initial ratings to MOS via weighted average. Following this stage, we propose the probability-based inference for LMMs to predict final scores.} 
    \vspace{-0.9em}
    \label{fig:human}
\end{figure*}

\paragraph{Video Quality Assessment (VQA).}

Named as video \textit{quality} assessment (VQA), the focus of this task is also kind of complicated, that several studies have claimed that scores are not only affected by quality issues, but also contents~\cite{vsfa}, and even aesthetics~\cite{wu2023dover}. Similar as IQA, while traditional approaches on VQA are typically based on handcraft features, \textit{e.g.} TLVQM~\cite{tlvqm}, VIDEVAL~\cite{videval}, and RAPIQUE~\cite{rapique}, recent deep-learning-based methods, such as VSFA~\cite{vsfa}, BVQA~\cite{bvqa2021}, DisCoVQA~\cite{discovqa} and SimpleVQA~\cite{simplevqa}, have shown much better performance and more robust OOD generalization. These efforts are further explored by FAST-VQA~\cite{fastvqa,wu2022fastervqa}, which proposes efficient end-to-end training to further advance VQA performance. Nevertheless, while the goal of VQA is similar to IQA (or IAA), the need to input \textit{videos} has hindered methods to tackle this task with the same modeling structure as image scoring approaches. A typical example is the CLIP-based attempts: as CLIP is image-based, though it can achieve good zero-shot VQA capabilities through a frame-by-frame inference~\cite{wu2023bvqi}, training CLIP-based methods on VQA datasets is extremely challenging~\cite{bvqiplus} and performs worse than specially-designed VQA models. In the proposed \textbf{\textsc{Q-Align}}, we utilize the language decoder to assemble videos as sequences of frames, so as to unify VQA with IQA/IAA under one structure, outperforming complicated specifically-designed architectures.


\paragraph{LMMs for Visual Scoring.}
Some recent investigations have discussed the possibilities for adopting Large Multi-modality Models (LMMs) for visual scoring. Namely, the Q-Bench~\cite{wu2023qbench} proposes a binary softmax strategy, enabling LMMs to predict quantifiable quality scores by extracting the softmax pooling result on logits of two frequent tokens (\textit{good/poor}). Based on this strategy, the Q-Instruct~\cite{wu2023qinstruct} notices that fine-tuning with text question-answering on related low-level queries can also improve visual scoring abilities of LMMs. Given insights from these studies, we design the \textbf{\textsc{Q-Align}} syllabus to systematically emulate the human rating and post-processing in visual scoring. Moreover, we demonstrate that the binary softmax strategy in Q-Bench is a simplified version equivalent to the collection process of MOS values from human ratings. Our experiments prove that with appropriate alignment strategies, LMMs can be more capable and robust visual scorers with \textit{the same (and even less) data used}.



\section{The \textsc{Q-Align}}

In this section, we elaborate on the proposed \textbf{\textsc{Q-Align}}. We start with our methodology to teach LMMs with rating levels (Sec.~\ref{sec:methodology}), and then discuss the proposed conversion strategy between rating levels and scores (Sec.~\ref{sec:conversion}). Afterwards, we introduce the unified structure (Sec.~\ref{sec:structure}) for images and videos, and conversation formats for each task (Sec.~\ref{sec:instructionformats}).

\subsection{Methodology}
\label{sec:methodology}

\subsubsection{[\textit{Insight 1}] How Do Humans Rate?}

To design the syllabus on training LMMs to score, we first review the process of collecting human opinions (Fig.~\ref{fig:human}). In general, the collection includes three stages as follows: 

\textbf{Stage 1: Training Human Raters.} As the standard process for collecting human opinions~\cite{itu}, the training process on human raters with the rating rules is vital, including aligning human raters with one or more examples for each \textbf{rating level} (Fig.~\ref{fig:human} \textit{left}, we take LSVQ~\cite{pvq} as an example). During this process, precise quality scores of the examples were \textit{not displayed} to human raters.

\textbf{Stage 2: Collecting human ratings.} After training human raters, the core stage is to collect initial human ratings (Fig.~\ref{fig:human} \textit{center}). In general, human raters may provide their opinions in two types: \textbf{1)} Directly choose rating levels. \textbf{2)} Toggle the slider to generate a score. In either way, human raters do not need to \textit{directly input} the scores to provide their opinions.

\textbf{Stage 3: Converting human ratings to MOS.} As in Fig.~\ref{fig:human} \textit{right}, initial ratings are averaged into MOS in visual scoring datasets. Human raters \textit{do not participate} in this stage.

During all three stages, human raters are \textbf{neither trained, nor instructed} to predict a score. This process is adopted because, in everyday life, when asked for an evaluation, people tend to respond with \textbf{qualitative adjectives} \textit{(for example, fine, poor, excellent)} rather than \textbf{numerical ratings} \textit{(e.g. 8.75, 1.08, 6.54)}. Thus, conducting the visual scoring tasks with rating levels utilizes this \textit{innate ability} of humans (\textit{providing qualitative adjectives}) to minimize their cognitive load, and improve the outcomes of subjective studies.

\subsubsection{[\textit{Insight 2}] How Do LMMs Rate?}

\begin{table}\small
    \vspace{-0.8em}
    \centering
    \renewcommand\arraystretch{1.14}
    \renewcommand\tabcolsep{4.5pt}
    \caption{[\textit{Insight 2}] \textsc{How do LMMs rate?} Responses of LMMs on  ``\textit{Rate the quality of the image}" from 1168 images in LIVE Challenge. LMMs prefer to respond with \textbf{qualitative adjectives.}}
    \resizebox{\linewidth}{!}{\begin{tabular}{l|ccc}
    \hline
     Model / Frequency & \textbf{Qualitative Adjectives} & \textbf{Numerical Ratings} \\  \hdashline
     Adapter-V2~\cite{llamaadapterv2} & \textbf{96\%} (1120/1168) & 4\% (48/1168) \\
     LLaVA-v1.5~\cite{improvedllava} & \textbf{100\%} (1168/1168) &  0\% (0/1168)\\
     mPLUG-Owl-2~\cite{mplugowl2} & \textbf{100\%} (1168/1168) & 0\% (0/1168)\\
     InstructBLIP~\cite{iblip} & \textbf{99\%} (1156/1168) & 1\% (12/1168) \\
     Shikra~\cite{shikra} & \textbf{100\%} (1168/1168) & 0\% (0/1168) \\
     \hline
    \end{tabular}}
    \vspace{-1.5em}
    \label{tab:lmminnate}
\end{table}

After analyzing the human opinion collection process, we further discover the ``\textit{innate ability}" of LMMs. Theoretically, fundamentally designed to understand and generate human-like text, LMMs should share similar behaviour patterns with humans. To validate this, we prompt five LMMs\footnote{None of them are explicitly trained for any visual rating tasks.} on the instruction as follows, and count their response statistics:

\textit{{\tt <img>} Rate the quality of the image.}

As results shown in Tab.~\ref{tab:lmminnate}, before specific alignment, LMMs predominantly respond with \textbf{qualitative adjectives}. Thus, if we use scores as the learning objective for LMMs, they need to first \textit{formally} learn to output scores, and then learn how to score accurately. To avoid this additional formatting cost, we choose rating levels instead as the targets of \textbf{\textsc{Q-Align}}.


\subsection{Conversion between Rating Levels and Scores}
\label{sec:conversion}

Based on the general methodology to teach LMMs with rating levels, we further discuss how to convert the scores in the existing datasets to discrete rating levels \textit{during training}, and how to obtain scores from LMMs \textit{during inference}. 

\subsubsection{[\textit{Training}] Scores $\to$ Rating levels.}

\paragraph{Equidistant Interval Partition.} During the training process, we convert the scores into discrete rating levels. Since adjacent levels in human rating are inherently equidistant (either Type 1 or Type 2, see Fig.~\ref{fig:human}), we also adopt equidistant intervals to convert scores into rating levels. Specifically, we uniformly divide the range between between the highest score ($\mathrm{M}$) and lowest score ($\mathrm{m}$) into five distinct intervals, and assign the scores in each interval as respective levels:
\begin{equation}
    {L(s)} = l_i \text{  if } \text{m} + \frac{i-1}{5}  \times \mathrm{(M-m)} < s \leq \mathrm{m} + \frac{i}{5} \times \mathrm{(M-m)}
\end{equation}
where \{$l_i|_{i=1}^{5}\}=\{\textit{bad, poor, fair, good, excellent}\}$ are the standard text rating levels as defined by ITU~\cite{itu}.

\begin{table}[htbp]
\vspace{-1em}
\centering
\small
\renewcommand\arraystretch{1.3}
\renewcommand\tabcolsep{3.5pt}
\caption{Precision of training conversion (\textit{Score $\to$ Rating Levels}) on the 5 training datasets for \textbf{\textsc{Q-Align}}. Metrics are SRCC/PLCC.}
\resizebox{\linewidth}{!}{\begin{tabular}{l:ccccc}
\hline
Conversion & \textbf{KonIQ} & \textbf{SPAQ} & \textbf{KADID} & \textbf{AVA} & \textbf{LSVQ} \\ \hdashline
\textit{Scores $\to$ Levels} & 0.952/0.961 & 0.969/0.968 & 0.979/0.982 & 0.920/0.930 & 0.940/0.944\\
\hline
\end{tabular}}
\label{tab:conversionprecision}
\vspace{-1.5em}
\end{table}

\paragraph{Precision of the Conversion.}~As the conversion mapping $L$ discussed above is a \textit{multi-to-one} mapping, it unavoidably slightly compromises the ground truth precision. In Tab.~\ref{tab:conversionprecision}, we record the conversion precision on the five datasets that we used for training \textbf{\textsc{Q-Align}}, that all conversion retains around \textbf{0.95} linear correlation (PLCC) with the scores. Considering that MOS values inherently have some randomness under such precision range, we believe that the converted rating levels are sufficiently accurate as training labels.

\subsubsection{[\textit{Inference}] Rating levels $\to$ Scores.}

After training, we need to convert the rating levels back to scores. Primarily, simulating the post-processing on human ratings (Fig.~\ref{fig:human} \textit{right}), we first define the reverse mapping $G$ from text-defined rating levels back to scores, as follows: 
\begin{equation}
    G: l_i \rightarrow i
\end{equation}
For instance, \textit{fair} is converted as score 3, and \textit{bad} as 1. 

For human opinion collection (Type 1), the MOS values are calculated via the weighted average of the converted scores and frequencies $f_{l_i}$ for each level: $\mathrm{MOS}=\sum_{i=1}^5 f_{l_i} G(l_i)$. For LMMs, we substitute the $f_{l_i}$ with the LMM-predicted probabilities for each rating level. Given that the predicted {\tt<LEVEL>} token of LMMs is the probability distribution (denoted as $\mathcal{X}$) on all possible tokens of the language model, we conduct a close-set softmax on \{$l_i|_{i=1}^{5}$\} to get the probabilities $p_{l_i}$ for each level, that $p_{l_i}$ for all $l_i$ sum as 1:

\begin{equation}
    p_{l_i} = \frac{e^{\mathcal{X}_{l_i}}}{\sum_{j=1}^{5} {e^{\mathcal{X}_{l_j}}}}
\end{equation}
and the final predicted scores of LMMs are denoted as 
\begin{equation}
    \mathrm{S_{LMM}}=\sum_{i=1}^5 p_{l_i} G(l_i) = i \times  \frac{e^{\mathcal{X}_{l_i}}}{\sum_{j=1}^{5} {e^{\mathcal{X}_{l_j}}}}
    \label{eq:5}
\end{equation}

The inference conversion is theoretically equivalent to the MOS collection process from a set of human ratings in levels. Moreover, it represents the general expression form of the binary softmax strategy ($\mathrm{S_\text{Q-Bench}}=\frac{e^{\mathcal{X}_\textit{good}}}{{e^{\mathcal{X}_\textit{good}}}+{e^{\mathcal{X}_\textit{poor}}}}$) as proposed by \citet{wu2023qbench}, which can be considered as a simplified version of Eq.~\ref{eq:5} with only two rating levels.

\begin{figure}
    \centering
    \includegraphics[width=\linewidth]{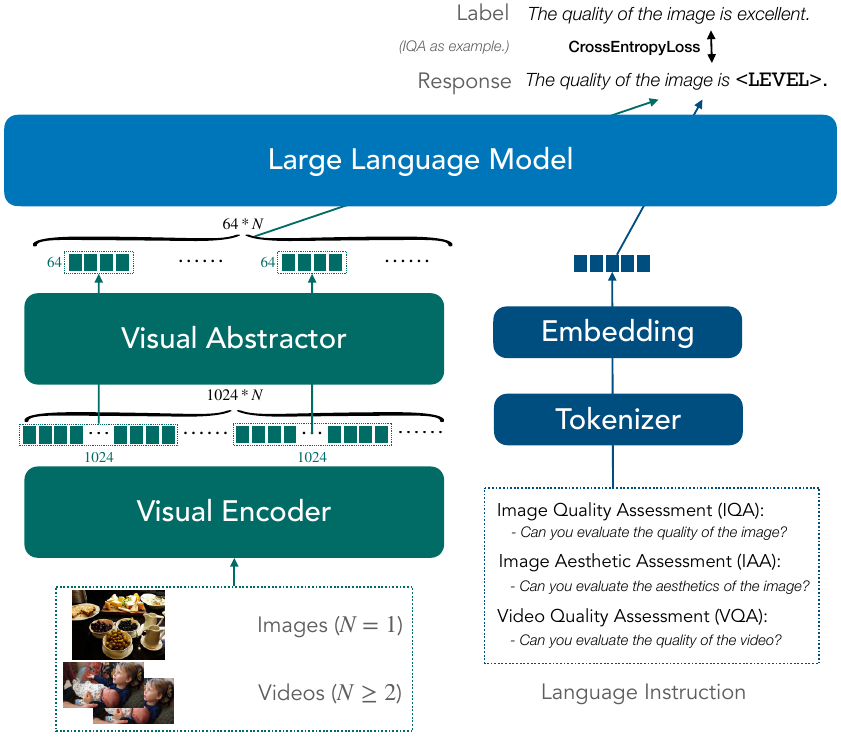}
    \vspace{-1.4em}
    \caption{Model structure of the \textbf{\textsc{Q-Align}}. By reducing tokens per image to 64 through the visual abstractor, it unifies images and videos (\textbf{as sequences of images}) under one general structure.}
    \label{fig:structure}
        \vspace{-1.3em}
\end{figure}

\subsection{Model Structure}
\label{sec:structure}

The model structure of the \textbf{\textsc{Q-Align}} (Fig.~\ref{fig:structure}) is based on the recently-published open-source LMM, mPLUG-Owl-2~\cite{mplugowl2}, which has proven exceptional visual perception ability as well as good language understanding ability. In the adopted structure, despite the visual encoder to convert images into embeddings, an additional visual abstractor further significantly reduces the token numbers per image ($1024\to64$). Under the $2048$ context length for LLaMA2~\cite{llama2}, we can feed as much as \textbf{30 images} (\textit{\textbf{2} without the abstractor})  together during supervised fine-tuning (SFT). This allows us to input a video \textbf{as a sequence of images} to LMM, and unify image (IQA, IAA) and video (VQA) scoring tasks under one structure. The \textbf{\textsc{Q-Align}} uses common GPT~\cite{gpt2} loss, \textit{i.e.} cross-entropy between labels and output logits.

\begin{table*}[htbp]
\vspace{-1em}
\centering
\small
\renewcommand\arraystretch{1.2}
\renewcommand\tabcolsep{8.8pt}
\caption{\textbf{\textsc{Q-Align}} and \textbf{\textsc{FewShot-Q-Align}} performance on image quality assessment (IQA). We adopt KonIQ and SPAQ (\textit{both in-the-wild photography}) as training set and evaluate on a wide range of test sets. The cross-set evaluations are labeled with $^\textsc{Cross}$.}
\resizebox{\linewidth}{!}{\begin{tabular}{ll:cccccccccc}
\hline
\textit{Training Set:}  \textbf{KonIQ$_\text{train}$} & \textit{$\to$Testing Set:} & \multicolumn{2}{c}{\textbf{KonIQ$_\text{test}$}} & \multicolumn{2}{c}{\textbf{SPAQ}$^\textsc{Cross}$} &\multicolumn{2}{c}{\textbf{LIVE Challenge}$^\textsc{Cross}$} &\multicolumn{2}{c}{\textbf{AGIQA-3K}$^\textsc{Cross}$} &\multicolumn{2}{c}{\textbf{KADID-10k}$^\textsc{Cross}$}  \\ \hline
\textbf{Method}  & \#Training &SRCC&PLCC & SRCC&PLCC & SRCC&PLCC & SRCC&PLCC & SRCC&PLCC \\
\hline
NIMA (TIP 2018)  & 7K (70\%) & 0.859 & 0.896 & 0.856 & 0.838 & 0.771 & 0.814 & 0.654 & 0.715 & 0.535 & 0.532 \\
DBCNN (TCSVT 2020) & 7K (70\%) & 0.875 & 0.884 & 0.806 & 0.812 & 0.755 & 0.773 & 0.641 & 0.730 &  0.484 & 0.497 \\
HyperIQA (CVPR 2020) & 7K (70\%)&   0.906 & 0.917 & 0.788 & 0.791 & 0.749 & 0.772 & 0.640 & 0.702 &0.468 & 0.506 \\
MUSIQ (ICCV 2021)  & 7K (70\%) & \underline{0.929} & \underline{0.924} & 0.863 & \underline{0.868} & 0.830 & 0.789 & 0.630 & 0.722 & 0.556 & 0.575\\
CLIP-IQA+ (AAAI 2023)  & 7K (70\%) & 0.895 & 0.909 &  0.864 & 0.866 & 0.805 & 0.832 & 0.685 & 0.736 &  0.654 & 0.653 \\
LIQE (CVPR 2023) & 7K (70\%) & {0.928} & 0.912 & 0.833 & 0.846 & \textbf{0.870} & 0.830 & 0.708 & \underline{0.772} &  \underline{0.662} & \underline{0.667} \\
\hline
\textbf{\textsc{FewShot-Q-Align}} (Ours) & \textbf{2K (20\%)} & 0.903 & 0.901 & \underline{0.871} & 0.860 & 0.840 & \underline{0.845} & \textbf{0.740} & \textbf{0.791} & 0.607 & 0.589  \\
\textbf{\textsc{Q-Align}} (Ours)  & 7K (70\%) & \textbf{0.940} & \textbf{0.941} & \textbf{0.887} & \textbf{0.886} & \underline{0.860} & \textbf{0.853} & \underline{0.735} & \underline{0.772} & \textbf{0.684} & \textbf{0.674}  \\
\hline
\end{tabular}}
\resizebox{\linewidth}{!}{\begin{tabular}{ll:cccccccccc}
\hline
\textit{Training Set:}  \textbf{SPAQ} &  \textit{$\to$Testing Set:} & \multicolumn{2}{c}{\textbf{KonIQ$_\text{test}$}$^\textsc{Cross}$} & \multicolumn{2}{c}{\textbf{SPAQ}} &\multicolumn{2}{c}{\textbf{LIVE Challenge}$^\textsc{Cross}$} &\multicolumn{2}{c}{\textbf{AGIQA-3K}$^\textsc{Cross}$} &\multicolumn{2}{c}{\textbf{KADID-10k}$^\textsc{Cross}$}  \\ \hline
\textbf{Method}  & \#Training &SRCC&PLCC & SRCC&PLCC & SRCC&PLCC & SRCC&PLCC & SRCC&PLCC \\
\hline
NIMA (TIP 2018)  & 8.8K (80\%) & 0.733 & 0.788 &  0.907 & 0.910 & 0.733 & 0.785 & 0.534 & 0.630 & 0.399 & 0.480 \\
DBCNN (TCSVT 2020) & 8.8K (80\%) & 0.731 & 0.758 & 0.908 & 0.913 & 0.702 & 0.748 & 0.459 & 0.518 & 0.490 & 0.508  \\
Fang \textit{et al.} (CVPR 2020) & 8.8K (80\%)& 0.714 & 0.742  &  0.908 &0.909 & 0.798 & 0.762& 0.570 & 0.649 & 0.381 & 0.448 \\
MUSIQ (ICCV 2021) & 8.8K (80\%)& 0.753 & 0.680 & 0.917 & \underline{0.921} & 0.813 & 0.789 & 0.564 & 0.675 & 0.349 & 0.429 \\
CLIP-IQA+ (AAAI 2023) & 8.8K (80\%)& 0.753 & 0.777 & 0.881 & 0.883 & 0.719 & 0.755 & 0.577 & 0.614 & 0.633 & 0.638 \\
LIQE (CVPR 2023) & 8.8K (80\%) & \underline{0.826} & \underline{0.847} & \underline{0.922} & 0.919 & 0.805 & \underline{0.866} & 0.672 & 0.722 & 0.639 & 0.627 \\
\hline
\textbf{\textsc{FewShot-Q-Align}} (Ours) & \textbf{2.2K (20\%)}   & 0.792 & 0.826 & 0.909 & 0.911 & \underline{0.823} & 0.834 & \underline{0.702} & \underline{0.772} & \underline{0.685} & \underline{0.678} \\
\textbf{\textsc{Q-Align}} (Ours) & 8.8K (80\%)& \textbf{0.848} & \textbf{0.879} & \textbf{0.930} & \textbf{0.933} &   \textbf{0.865} & \textbf{0.873} & \textbf{0.723} & \textbf{0.786} & \textbf{0.743} & \textbf{0.740} \\
\hline
\end{tabular}}
\label{tab:iqa}
\vspace{-1.2em}
\end{table*}

\subsection{Conversation Formats}
\label{sec:instructionformats}

In this section, we define the conversation formats for each task. Denote the image token as {\tt <img>}, the converted level for the image or video as {\tt <level>}, the exemplar conversation formats for each task are as follows:

\textbf{Image Quality Assessment (IQA)} \\
\textit{\#User:} {\tt <img>} \textit{Can you evaluate the quality of the image?} \\
\textit{\#Assistant:} \textit{The quality of the image is {\tt <level>}.}

\textbf{Image Aesthetic Assessment (IAA)} \\
\textit{\#User:} {\tt <img>} \textit{How is the aesthetics of the image?} \\
\textit{\#Assistant:} \textit{The aesthetics of the image is {\tt <level>}.}

\textbf{Video Quality Assessment (VQA)} \\
\textit{\#User:} {\tt <img>} \textit{Rate the quality of the video.} \\
\textit{\#Assistant:} \textit{The quality of the video is {\tt <level>}.}

The user queries are randomly chosen from a group of \textit{paraphrases} as an augmentation. Following \citet{vicuna}, only the LMM responses (after \textit{\#Assistant:}) are supervised.


\section{Experiments}

\begin{table*}[htbp]
\centering
\small
\renewcommand\arraystretch{1.18}
\renewcommand\tabcolsep{5pt}
\caption{\textsc{Mix-Data} experiments for \textbf{\textsc{Q-Align}} on image quality assessment (IQA). We label intra-dataset testing sets for each training set combination with gray background, with rest as cross-set settings. Mixing datasets notably improves unseen dataset performance on IQA.}
\resizebox{\linewidth}{!}{\begin{tabular}{l:cccccccccccccc}
\hline
{\textit{Testing Set:}} & \multicolumn{2}{c}{\textbf{KonIQ$_\text{test}$}} & \multicolumn{2}{c}{\textbf{SPAQ}} &\multicolumn{2}{c}{\textbf{KADID-10k}}  &\multicolumn{2}{c}{\textbf{LIVE Challenge}} &\multicolumn{2}{c}{\textbf{AGIQA-3K}} &\multicolumn{2}{c}{\textbf{LIVE}} &\multicolumn{2}{c}{\textbf{CSIQ}}  \\ \hline
 \textit{Training Set:}  &SRCC&PLCC & SRCC&PLCC & SRCC&PLCC & SRCC&PLCC & SRCC&PLCC & SRCC&PLCC & SRCC&PLCC \\
\hline
 \textit{KonIQ} & \colorbox{lightgray}{\textbf{0.940}} & \colorbox{lightgray}{0.941} & {0.887} & {0.886} & {0.684} & {0.674}  & {0.860} & {0.853} & {0.735} & {0.772}  & 0.867 & 0.838 & 0.700 & 0.759   \\
 \textit{SPAQ} & {0.848} & {0.879} & \colorbox{lightgray}{0.930} & \colorbox{lightgray}{0.933} & {0.743} & 0.740 &   0.865 & 0.873 & 0.723 & 0.786 & 0.861 & 0.822 & 0.733 & 0.781\\
 \textit{KonIQ} + \textit{SPAQ} & \colorbox{lightgray}{\textbf{0.940}} & \colorbox{lightgray}{0.943} &  \colorbox{lightgray}{\textbf{0.931}} & \colorbox{lightgray}{\textbf{0.933}} & 0.708 & 0.692 & 0.879 & 0.883 & 0.727 & \textbf{0.795} & 0.859 & 0.827 & 0.767 & 0.795 \\
 \textit{KADID} & 0.668 & 0.665 & 0.860 &  0.854 & \colorbox{lightgray}{0.919} & \colorbox{lightgray}{0.918} & 0.702 &  0.744 & 0.711 & 0.712 & 0.809 & 0.791 & 0.756 & 0.784   \\

\textit{KonIQ} + \textit{SPAQ} + \textit{KADID} & \colorbox{lightgray}{0.938} & \colorbox{lightgray}{\textbf{0.945}} &  \colorbox{lightgray}{\textbf{0.931}} & \colorbox{lightgray}{\textbf{0.933}} &  \colorbox{lightgray}{\textbf{0.934}} & \colorbox{lightgray}{\textbf{0.935}} & \textbf{0.883} & \textbf{0.887} & \textbf{0.733} & 0.788  & \textbf{0.870} & \textbf{0.840} & \textbf{0.845} & \textbf{0.876}  \\
\hline
\end{tabular}}
\vspace{-1.5em}
\label{tab:mix}
\end{table*}

\begin{table*}[htbp]
\centering
\small
\renewcommand\arraystretch{1.18}
\renewcommand\tabcolsep{7.5pt}
\caption{\textbf{\textsc{Q-Align}} performance on video quality assessment (VQA). All methods are trained on the same dataset (LSVQ$_\text{train}$) and evaluated on two intra-dataset (LSVQ$_\text{test}$ and LSVQ$_\text{1080p}$) and two cross-dataset (KoNViD-1k and MaxWell$_\text{test}$) test sets.}
\resizebox{\linewidth}{!}{\begin{tabular}{l:ccccccccc}
\hline
\textit{Training Set:}  \textbf{LSVQ$_\text{train}$} & $\to$\textit{Testing Set:} & \multicolumn{2}{c}{\textbf{LSVQ$_\text{test}$}} & \multicolumn{2}{c}{\textbf{LSVQ$_\text{1080p}$}} & \multicolumn{2}{c}{\textbf{KoNViD-1k}$^\textsc{Cross}$} & \multicolumn{2}{c}{\textbf{MaxWell$_\text{test}$}$^\textsc{Cross}$} \\ \hline
\textbf{Method} & IQA Pre-training? & SRCC&PLCC& SRCC&PLCC& SRCC&PLCC& SRCC&PLCC  \\
\hline
TLVQM (TIP 2019) & \xmark & 0.772&0.774&0.589&0.616&0.732&0.724& -- & -- \\
VSFA (ACMMM 2019) & \xmark & 0.801&0.796&0.675&0.704&0.784&0.794& -- & -- \\
VIDEVAL (TIP 2021) & \xmark & 0.794&0.783&0.545&0.554&0.751&0.741& -- & -- \\
PVQ (CVPR 2021) & \cmark & 0.827 & 0.828 & 0.711 & 0.739 & 0.791 & 0.795 & 0.618 & 0.634  \\
BVQA (TCSVT 2022) & \cmark & 0.852&0.854 & 0.772&0.788 & 0.839&0.830 & 0.675 & 0.673 \\
DisCoVQA (TCSVT 2023) & \xmark & 0.859&0.850 & 0.734&0.772 & 0.851&0.853 & 0.704&0.687 \\
SimpleVQA (ACMMM 2022) & \cmark & 0.867&0.861 & 0.764&0.803 & 0.840&0.834 & 0.720&0.715 \\
FAST-VQA (ECCV 2022) & \xmark & 0.876&0.877 & 0.779&0.814 & 0.859&0.855 & 0.720&0.728 \\ \hline
\textbf{\textsc{Q-Align} (Ours)} (1fps) & \xmark & \textbf{0.883}&\textbf{0.882} & \textbf{0.797}&\textbf{0.830} & \textbf{0.865}&\textbf{0.877} & \textbf{0.780}&\textbf{0.782} \\ \hline
\multicolumn{10}{l}{\textit{--- Ensemble-based Approaches}} \\  \hdashline
DOVER (\textit{aesthetic branch} + FAST-VQA, ICCV 2023) & \xmark & 0.886 & 0.887 & 0.795 & 0.830 & 0.883 & 0.884 & 0.748 & 0.755 \\
\textbf{\textsc{Q-Align} (Ours)} (1fps) + FAST-VQA & \xmark & \textbf{0.899} & \textbf{0.899} & \textbf{0.818} & \textbf{0.850} & \textbf{0.895} & \textbf{0.897} &  \textbf{0.779} & \textbf{0.784}\\ \hline
\end{tabular}}
\label{tab:vqa}
\vspace{-1.5em}
\end{table*}

\begin{table}[htbp]
\centering
\small
\renewcommand\arraystretch{1.18}
\renewcommand\tabcolsep{5pt}
\caption{\textbf{\textsc{Q-Align}} performance on image aesthetic assessment (IAA). All methods are trained under the \textsc{official} split setting.}
\resizebox{\linewidth}{!}{\begin{tabular}{ll:ccc}
\hline
\multicolumn{2}{l}{\textit{Training Set:}  \textbf{AVA$_\text{train}$}} & $\to$\textit{Testing Set:} & \multicolumn{2}{c}{\textbf{AVA$_\text{test}$}}  \\ \hline
\textbf{Method} & \#Training & Extra Data? & SRCC&PLCC  \\
\hline
NIMA (TIP 2018) & 236K (92\%) & \xmark & 0.612 & 0.636 \\
MLSP (CVPR 2019) &  236K (92\%) &  \xmark & 0.756 & 0.757 \\
MUSIQ (ICCV 2021) & 236K (92\%) & \xmark & 0.726 & 0.738 \\
MaxViT (ECCV 2022) & 236K (92\%)& \xmark & 0.708 & 0.745 \\
CLIP-IQA+ (AAAI 2023)& 236K (92\%) & \xmark & 0.619 & 0.586 \\
Aesthetic Predictor (2023) & 236K (92\%)& \xmark & 0.721 & 0.723 \\
LIQE (CVPR 2023) & 236K (92\%) & \xmark & 0.776 & 0.763\\
VILA (CVPR 2023)  & 236K (92\%)& \cmark & 0.774 & 0.774 \\
\hline
\textbf{\textsc{FewShot-Q-Align}} (Ours) & \textbf{26K (10\%)}  &  \xmark & 0.776 & 0.775\\
\textbf{\textsc{Q-Align}} (Ours) & 236K (92\%) &  \xmark & \textbf{0.822} & \textbf{0.817} \\
\hline
\end{tabular}}
\vspace{-1.5em}
\label{tab:iaa}
\end{table}

\begin{table*}[htbp]
\centering
\small
\renewcommand\arraystretch{1.25}
\renewcommand\tabcolsep{3.5pt}
\caption{Results of \textbf{\textsc{OneAlign}} as one unified model for IQA, IAA and VQA, in comparison with single task experts (IQA, IAA, VQA) and partly multi-task experts (IQA+IAA, IQA+VQA, IAA+VQA). LIVE-C abbreviates for LIVE Challenge. Metrics are SRCC/PLCC.}
\resizebox{\linewidth}{!}{\begin{tabular}{l:ccccccc:c:cccc}
\hline
{\textit{Training / Testing Set}} & \textbf{KonIQ} & \textbf{SPAQ} & \textbf{KADID} & \textbf{LIVE-C} & \textbf{AGIQA} & \textbf{LIVE} & \textbf{CSIQ} & \textbf{AVA} & \textbf{LSVQ$_\text{test}$} & \textbf{LSVQ$_\text{1080P}$} & \textbf{KoNViD} & \textbf{MaxWell}  \\ \hline
{IQA}$^\textit{(KonIQ + SPAQ + KADID)}$ & \colorbox{lightgray}{.938/.945} &  \colorbox{lightgray}{.931/.933} &  \colorbox{lightgray}{{.934}/{.935}} & {.883}/{.887} & {.733}/.788  & .870/.840 & .845/.876 & .208/.228 & .755/.757 & .680/.718 & .799/.806 & .682/.694 \\
{VQA}$^\textit{(LSVQ)}$ & .731/.788 & .841/.819 & .659/.651 & .715/.727 & .780/.834 & .826/.797 & .755/.814 & .289/.323 & \colorbox{lightgray}{.883/.882} & \colorbox{lightgray}{.797/.830} & .865/.877 & .780/.782 \\
{IAA}$^\textit{(AVA)}$ & .574/.603 & .662/.653 & .536/.547 & .685/.636 & .750/.792 &  .770/.740 & .527/.596 & \colorbox{lightgray}{.822/.817} & .624/.600 & .515/.511 & .717/.681 & .659/.648 \\ \hdashline
{IQA + VQA} & \colorbox{lightgray}{.944/.949} & 	\colorbox{lightgray}{.931/.934} &	\colorbox{lightgray}{.952/.953} &	.892/.899 & .739/.782 & .874/.846 & .852/.876 & .197/.222  & 	\colorbox{lightgray}{.885/.883}	 & \colorbox{lightgray}{.802/.829}	& .867/.880	& .781/.787  \\
{IQA + IAA} & \colorbox{lightgray}{.940/.947} & \colorbox{lightgray}{.931/.933} & \colorbox{lightgray}{.945/.945} & {.862/.868} & .782/.824 & .895/.864 & .865/.883 & \colorbox{lightgray}{.822/.819} & .785/.785 & .700/.730 & .831/.829  & .716/.728 \\
{IAA + VQA} & .640/.664 & .740/.732 & .626/.632 & .703/.669 & .769/.819 & .794/.769 & .558/.628 & \colorbox{lightgray}{.822/.819} & \colorbox{lightgray}{.886/.885} & \colorbox{lightgray}{.800/.834} & .874/.884 & .776/.781   \\ \hdashline
All (\textbf{\textsc{OneAlign}}) & \colorbox{lightgray}{.941/.950}	& \colorbox{lightgray}{.932/.935} &	\colorbox{lightgray}{.941/.942} &	.881/.894 &	.801/.838 & .887/.856 &	.881/.906 & \colorbox{lightgray}{.823/.819} & \colorbox{lightgray}{.886/.886} & \colorbox{lightgray}{.803/.837} & .876/.888 &.781/.786\\
\hline
\end{tabular}}
\label{tab:onealign}
\vspace{-1.1em}
\end{table*}

\subsection{Experimental Settings}

In experiments, we fine-tune from the pre-trained weights of mPLUG-Owl2~\cite{mplugowl2}. We set batch sizes as 64 for all IQA/VQA datasets, and 128 while the \textbf{AVA} dataset is involved in training. The learning rate is set as $2e-5$, and we train for 2 epochs for all, except for few-shot settings, where we train for 4 epochs to make the models fully converge. The same as \citet{improvedllava}, all reported performance of \textbf{\textsc{Q-Align}} are evaluated on the final weights after training. We conduct training on 4*NVIDIA A100 80G, and report inference latency on one RTX3090 24G GPU. For videos, we sparsely sample one frame per second (\textit{1fps}).

\subsection{Datasets}

\textbf{IQA datasets.} We choose the {KonIQ-10k} (\textit{in-the-wild}), {SPAQ} (11K, \textit{in-the-wild}), and {KADID-10k} (\textit{synthetic}) as training sets to train the \textbf{\textsc{Q-Align}} on IQA. Despite evaluating on the test sets on the three training datasets, we also evaluate on four unseen datasets: {LIVE Challenge} (1.1K, \textit{in-the-wild}), {AGIQA-3K} ({AI-generated}), {LIVE} and {CSIQ} (\textit{both synthetic}) to examine its OOD generalization ability.

\textbf{IAA datasets.} We choose the well-recognized {AVA}~\cite{ava} dataset to compare aesthetic abilities between the \textbf{\textsc{Q-Align}} and existing approaches. Following \citet{clipiaa}, we conduct experiments on the \textsc{official} train-test split with \textbf{236K} training images and 19K test images.

\textbf{VQA datasets.} We choose the largest in-the-wild VQA dataset, {LSVQ}, with 28K training videos to train the \textbf{\textsc{Q-Align}} on VQA. Similar as IQA, we test on two official test sets of LSVQ ({LSVQ$_\text{test}$} and {LSVQ$_\text{1080P}$}), and two unseen datasets, {KoNViD-1k} and {MaxWell} for OOD evaluation.

\subsection{Results on Individual Tasks}

\subsubsection{Image Quality Assessment (IQA)}

For IQA, we first compare the conventional setting where models are trained on a single dataset. As shown in Tab.~\ref{tab:iqa}, while CLIP-based methods (CLIP-IQA+ and LIQE) show only comparable or even worse performance on intra-dataset settings than the visual-only state-of-the-art, MUSIQ, the proposed \textbf{\textsc{Q-Align}} can notably achieve better accuracy than all visual-only approaches. On cross-dataset settings (OOD generalization), \textbf{\textsc{Q-Align}} significantly improves visual-only methods by more than \textbf{10\%}, and CLIP-IQA+ and LIQE by \textbf{8\%} and \textbf{4\%} respectively. In summary, LMM-based \textbf{\textsc{Q-Align}} is more competitive under the same data.

In addition to the conventional setting, we further validate that the \textbf{\textsc{Q-Align}} can achieve high accuracy with \textit{even less data}. Denoted as \textbf{\textsc{FewShot-Q-Align}} in Tab.~\ref{tab:iqa}, it can reach comparable performance with existing state-of-the-art IQA approaches by using only 20\% images in the datasets, suggesting that the proposed rating-level based approach can swiftly activate LMM's inherent knowledge about IQA.

Despite single dataset results, we further evaluate the mix-data scenario for \textbf{\textsc{Q-Align}}, as listed in Tab.~\ref{tab:mix}. In summary, while traditional IQA methods~\cite{dbcnn,musiq,liqe} have reported to experience reduced accuracy while mixing in-the-wild (KonIQ, SPAQ) and synthetic (KADID) IQA datasets for training, we demonstrate that via the simple mixing strategy (concatenate datasets), the \textbf{\textsc{Q-Align}} is able to retain or improve the accuracy on individual datasets while mixing datasets. This ability paves the way for the ultimate \textbf{\textsc{OneAlign}} that unifies and mixes datasets from different scoring tasks. 

\subsubsection{Image Aesthetic Assessment (IAA)}

In Tab.~\ref{tab:iaa}, we list the results of the \textbf{\textsc{Q-Align}} and existing state-of-the-arts on IAA. Compared with IQA, IAA is much more complicated, and the \textbf{\textsc{Q-Align}} exhibits far larger advantages with its larger model capacity. It can outperform LIQE by \textbf{7\%}, Aesthetic Predictor~\cite{aestheticpredictor} by \textbf{10\%}. It even significantly improves VILA, which is additionally pre-trained by AVA-Captions, by a notable \textbf{6\%} margin. Moreover, similar as IQA, the \textbf{\textsc{FewShot-Q-Align}} is able to outperform existing IAA methods with only 10\% of AVA dataset used for training, further proving the data efficiency of the proposed syllabus on aligning LMMs for scoring.

\subsubsection{Video Quality Assessment (VQA)}

As listed in Tab.~\ref{tab:vqa}, with only sparse frames (\textit{1fps}) as inputs, the \textbf{\textsc{Q-Align}} is able to outperform specially-designed VQA approaches with complicated temporal modules and all frames fed into their models. Similar as IQA, it exhibits excellent OOD generalization and surpasses FAST-VQA by 6\% on cross-dataset evaluation from LSVQ$_\text{train}$ to MaxWell$_\text{test}$ dataset. While \textbf{\textsc{Q-Align}} alone can already reach comparable accuracy with DOVER, an approach that ensembles a sparse-frame aesthetic branch with FAST-VQA, it can similarly achieve better performance in this ensemble setting. The ensemble of \textbf{\textsc{Q-Align}} and FAST-VQA proves over 1\% advantage to DOVER on all four evaluation datasets. On the other hand, the ensemble gain with FAST-VQA suggests that though the \textbf{\textsc{Q-Align}} (1fps) has been effective, since we haven't yet input all frames of the video into the LMM, there is still room for future improvements.

\subsection{The \textsc{OneAlign}}

Previous evaluations have revealed two exciting abilities of \textbf{\textsc{Q-Align}}. First, it reaches state-of-the-art with notable improvements on IQA, IAA, and VQA under one unified structure. Second, it shows good mix-dataset learning capacity. Given the abilities, we further combine all training datasets for the three tasks to train the \textbf{\textsc{OneAlign}}, the all-in-one visual scorer. As evaluated in Tab.~\ref{tab:onealign}, all multi-task variants have shown improved performance than single-task variants. Moreover, the \textbf{\textsc{OneAlign}} remarkably improves OOD generalization on several unseen datasets: AGIQA$^{+6.8\%\text{SRCC}}$, CSIQ$^{+3.6\%\text{SRCC}}$, LIVE$^{+1.7\%\text{SRCC}}$, KoNViD$^{+1.1\%\text{SRCC}}$. We hope that the \textbf{\textsc{OneAlign}} can be widely applied to real-world scenarios, pioneering the paradigm shift in this field.

\subsection{Cost Analysis}

\subsubsection{Training Cost}

\begin{table}[htbp]
\vspace{-1em}
\centering
\small
\renewcommand\arraystretch{1.2}
\renewcommand\tabcolsep{4pt}
\caption{Epochs required to converge to best results for IQA on KonIQ-10k dataset. Metrics are the average of SRCC and PLCC.}
\resizebox{\linewidth}{!}{\begin{tabular}{l:cccc}
\hline
\textit{Method} & \textit{best} ($\uparrow$) & \textit{Ep1} ($\uparrow$) & \textit{Ep1} - \textit{best} ($\uparrow$)& \#Epochs for \textit{best} ($\downarrow$)\\ \hline
NIMA (TIP 2018) & 0.870 & 0.650 & -0.220 & 15\\
CLIP-IQA+ (AAAI 2023) & 0.903 & 0.825 & -0.078 & 12 \\
LIQE (CVPR 2023) & 0.920 & 0.887  & -0.033 & 9 \\ \hdashline
 \textbf{\textsc{Q-Align}} (Ours)  & \textbf{0.942} & \textbf{0.931} &  \textbf{-0.011} & \textbf{2}\\
\hline
\end{tabular}}
\label{tab:trainingspeed}
\vspace{-1.2em}
\end{table}

As compared in Tab.~\ref{tab:trainingspeed}, the \textbf{\textsc{Q-Align}} can converge with fewer iterations than existing IQA methods ($bs=64$ for all), including CLIP-based methods. While existing methods usually need about 10 epochs to reach the best result, the \textbf{\textsc{Q-Align}} can outperform all existing methods with only one epoch, and obtain the best results in 2 epochs. With 4*A100 80G GPU, it requires only \textbf{9 minutes to converge}, which costs $<2$ USD from most cloud GPU providers.

\begin{table}[htbp]
\vspace{-1em}
\centering
\small
\renewcommand\arraystretch{1.2}
\renewcommand\tabcolsep{3.3pt}
\caption{Inference latency and throughput of the \textbf{\textsc{Q-Align}} on \textit{images} on RTX3090. Larger batch sizes ($>$64) will cause OOM.}
\resizebox{\linewidth}{!}{\begin{tabular}{l:ccccccc}
\hline
Batch Size & 1 & 2 & 4 & 8 & 16 & 32 & 64 \\
\hline
\textbf{Latency} (\textit{ms}) & 101 & 154 & 239 & 414 & 757 & 1441 & 2790 \\
\textbf{Throughput} (\textit{image/sec}) & 9.90 & 12.99 & 16.74 & 19.32 & 21.14 & 22.21 & 22.94 \\
\hline
\end{tabular}}
\label{tab:inferencecost}
\vspace{-1em}
\caption{Latency and throughput on \textit{videos}. As videos have variable lengths, we set batch size as 1 for them to avoid padding cost.}
\renewcommand\tabcolsep{6.2pt}
\resizebox{\linewidth}{!}{\begin{tabular}{l:ccccccc}
\hline
Video Length (\textit{sec}) &  5 & 7 & 8 & 9 & 10 & 11 & 12 \\
\hline
\textbf{Latency} (\textit{ms}) & 236 & 315 & 350 &  377 &  430 & 463 & 514 \\
\textbf{Throughput} (\textit{video/sec}) & 4.24 & 3.17 & 2.86 &  2.65 & 2.33 &  2.16 & 1.95  \\
\hline
\end{tabular}}
\label{tab:inferencecostvid}
\end{table}

\subsubsection{Inference Latency}

In Tab.~\ref{tab:inferencecost} and Tab.~\ref{tab:inferencecostvid}, we discuss the inference latency of \textbf{\textsc{Q-Align}} on images and videos. With the best GPU utilization, in one second, it can predict scores on \textbf{23} images, \textbf{4.2} 5s-duration videos, or \textbf{1.9} 12s-duration videos on a single consumer-level RTX3090 GPU. Achieving $20\times$ faster than real-time on videos, its low inference latency allows wider real-world applications of the LMM-based visual scorer.

\begin{table}[htbp]
\centering
\small
\renewcommand\arraystretch{1.18}
\renewcommand\tabcolsep{3pt}
\vspace{-1.5em}
\caption{\textbf{\textsc{Q-Align}} compared with the variant that use scores (\textit{in {\tt.2f} format}) as \textit{training} objective. Metrics are (SRCC+PLCC)/2.}
\resizebox{\linewidth}{!}{\begin{tabular}{l:ccccc}
\hline
\textit{Training Set:} & \multicolumn{5}{c:}{  \textbf{KonIQ}}  \\ \hdashline
\textit{Testing Set:} & \textbf{KonIQ} & \textbf{SPAQ}$^\textsc{Cross}$ & \textbf{LIVE-C}$^\textsc{Cross}$ & \textbf{AGIQA}$^\textsc{Cross}$ & \textbf{KADID}$^\textsc{Cross}$ \\\hline
Existing SOTA & \underline{0.926} & \underline{0.865} & \underline{0.850} & \underline{0.740} & \underline{0.665} \\
- \textit{Training with Scores} & 0.921 & 0.858 & 0.793 & 0.731 & 0.524 \\
\textbf{\textsc{Q-Align}} (Ours) & \textbf{0.941} & \textbf{0.887}& \textbf{0.857} & \textbf{0.754} & \textbf{0.679} \\ \hdashline
Levels \textbf{\textit{vs}} Scores & \textbf{+2.2\%} & \textbf{+3.4\%} & \textbf{+8.1\%} & \textbf{+3.1\%} &  \textbf{+29.6\%}  \\
\hline
\end{tabular}}
\resizebox{\linewidth}{!}{\begin{tabular}{l:ccccc}
\hline
\textit{Training Set:} &\multicolumn{5}{c}{\textbf{SPAQ}}  \\ \hdashline
\textit{Testing Set:} &  \textbf{SPAQ} & \textbf{KonIQ}$^\textsc{Cross}$ & \textbf{LIVE-C}$^\textsc{Cross}$ & \textbf{AGIQA}$^\textsc{Cross}$ & \textbf{KADID}$^\textsc{Cross}$\\\hline
Existing SOTA & \underline{0.921} & \underline{0.836} & \underline{0.835} & \underline{0.697} & \underline{0.633}  \\
- \textit{Training with Scores} & 0.918 & 0.813 & 0.813 & 0.657 & 0.485\\
\textbf{\textsc{Q-Align}} (Ours) & \textbf{0.932} & \textbf{0.863} & \textbf{0.869} & \textbf{0.755} & \textbf{0.741}\\ \hdashline
Levels \textbf{\textit{vs}} Scores & \textbf{+1.5\%} & \textbf{+6.2\%} & \textbf{+6.9\%} & \textbf{14.9\%} &\textbf{+52.8\%}\\
\hline
\end{tabular}}
\vspace{-1.5em}
\label{tab:levelvsscore}
\end{table}

\subsection{Ablation Studies}

\textbf{\textbf{\textsc{Q-Align}} \textit{vs} training with scores.} In Tab.~\ref{tab:levelvsscore}, we compare the \textbf{\textsc{Q-Align}} with the variant that directly instructs the LMM to output scores during training. Using the proposed level-based syllabus can lead to in-average \textbf{10\%}  improvements on cross-dataset (OOD) evaluations than the score-based syllabus, suggesting that the proposed syllabus better inherits the innate visual judgment abilities of LMMs. It especially achieves 40\% gain over the score-based variant from SPAQ/KonIQ (\textit{in-the-wild}) to KADID (\textit{synthetic}), further proving that the \textbf{\textsc{Q-Align}} better activates the original knowledge of LMMs.
Moreover, we demonstrate that, the accuracies of score-based alignment cannot even surpass existing state-of-the-art on any settings, wasting the large capacities of these powerful foundation models.



\begin{table}[!t]
\centering
\renewcommand\arraystretch{1.08}
\renewcommand\tabcolsep{3.6pt}
\caption{\textbf{\textsc{OneAlign}} predictions on real-world images, from logits to probabilities, and finally to scores. View more in Appendix. }
\resizebox{\linewidth}{!}{\begin{tabular}{l:ccccc:ccccc}
\hline
{} & \multicolumn{5}{c:}{(A)} & \multicolumn{5}{c}{(B)} \\ \hline
{\tt <img>} & \multicolumn{5}{c:}{\includegraphics[width=5cm]{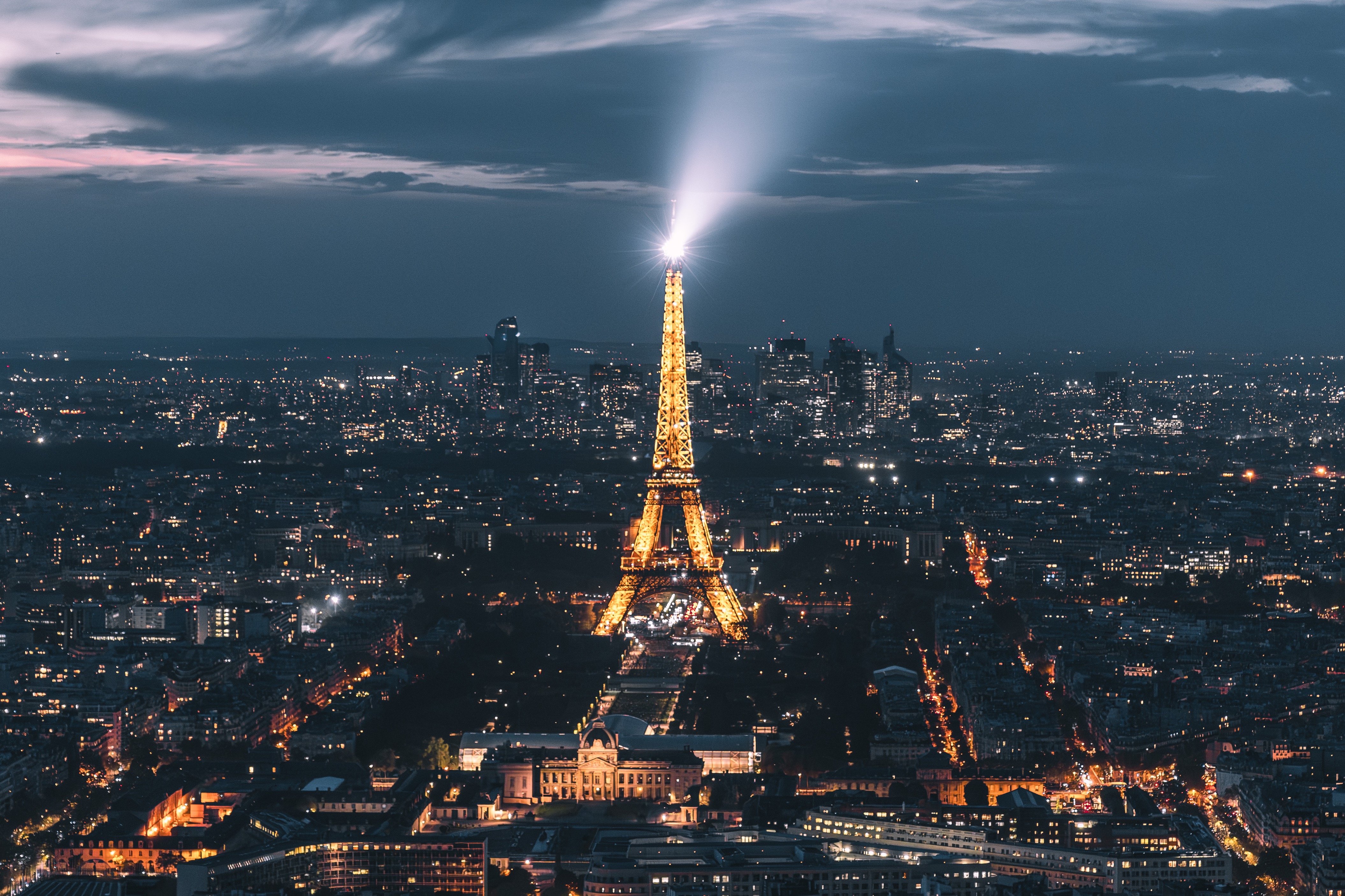}} & \multicolumn{5}{c}{\includegraphics[width=5cm]{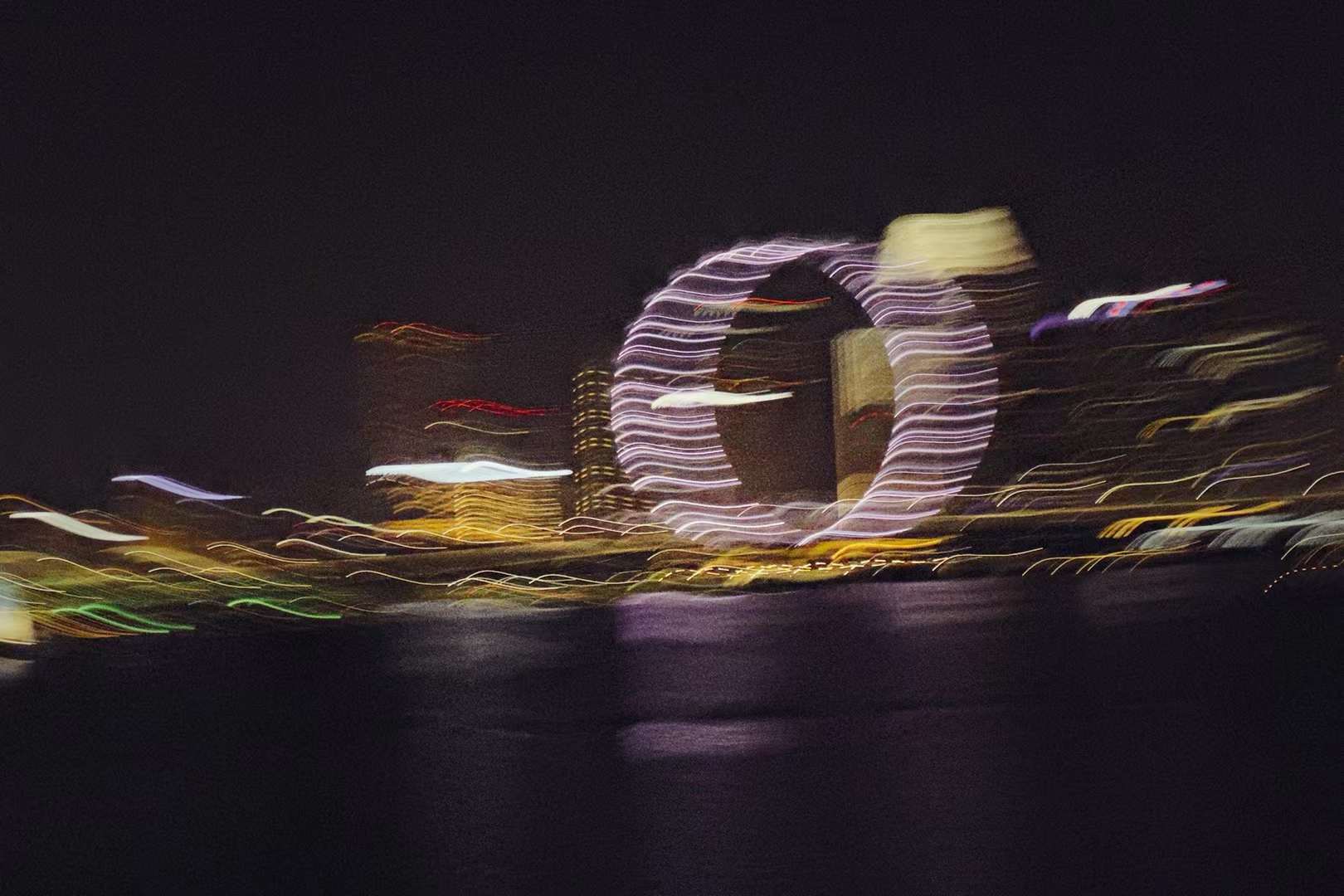}}  \\ \hline
$l_i$ & \textit{excellent} & \textit{good} & \textit{fair} & \textit{poor} & \textit{bad} & \textit{excellent} & \textit{good} & \textit{fair} & \textit{poor} & \textit{bad}\\ \hdashline
$\mathcal{X}^{l_i}_\text{IQA}$ & \underline{18.03} & \textbf{18.38} & 14.63 &  11.60 & 9.477 & 8.953 & 11.37 & 15.31 & \textbf{18.06} & \underline{16.59}   \\
${p}^{l_i}_\text{IQA}$ & \underline{0.409} & \textbf{0.577} & 0.014 &  0.000 & 0.000 & 0.000 & 0.001 & 0.050 & \textbf{0.772} & \underline{0.178} \\
$S_\text{LMM, IQA}$ &  \multicolumn{5}{c:}{\textbf{4.3926} \textcolor{gray}{(Range: [1,5])}}& \multicolumn{5}{c}{1.8740 \textcolor{gray}{(Range: [1,5])}} \\ \hdashline
$\mathcal{X}^{l_i}_\text{IAA}$ & \underline{16.63} & \textbf{18.17} & 15.77 &  12.13 & 10.77 &9.594 &  13.13 & \underline{16.95} &  \textbf{17.67} &  14.91  \\
${p}^{l_i}_\text{IAA}$ & \underline{0.163} &  \textbf{0.766} & 0.069 & 0.002 & 0.000  & 0.000 & 0.007 & \underline{0.312} &  \textbf{0.641} &  0.040\\
$S_\text{LMM, IAA}$ & \multicolumn{5}{c:}{\textbf{4.0879} \textcolor{gray}{(Range: [1,5])}}& \multicolumn{5}{c}{2.2861 \textcolor{gray}{(Range: [1,5])}}   \\
\hline
\end{tabular}}
\vspace{-1em}
\label{tab:qual}
\end{table}
\subsection{Qualitative Analysis}

In Tab.~\ref{tab:qual}, we visualize the IQA and IAA prediction results of the \textbf{\textsc{OneAlign}} on two real-world images. Despite the basic ability to judge that (A) $>$ (B) in both \textit{quality} and \textit{aesthetics}, we notice that it can further capture subtle differences. Though trained with only discrete levels, its \textit{2nd highest level} (\underline{underlined}) can provide finer-grained evaluations, that the \textit{aesthetics} of (B) is between \textit{fair} and \textit{poor}, while its \textit{quality} lies between \textit{poor} and \textit{bad}. Moreover, though the levels are not explicitly ordinal during training, we find out that it never predicts \textit{1st} and \textit{2nd} highest levels on non-adjacent ratings (\textit{e.g. good\&poor}), suggesting that LMMs inherently understand these text-defined ratings.

\section{Conclusion}
In conclusion, our paper marks a significant stride in the realm of visual scoring by innovatively instructing Large Multi-modality Models (LMMs) with discrete text-defined levels (\textit{e.g., good, poor}) rather than direct scores (\textit{e.g., 3.45, 1.77}). This syllabus, named the \textbf{\textsc{Q-Align}}, has achieved remarkable improvements over state-of-the-art IQA, IAA and VQA approaches under one general structure, and further unifies all the three tasks under one single model, the \textbf{\textsc{OneAlign}}. The \textbf{\textsc{Q-Align}} unlocks the potential of LMMs in predicting accurate and robust visual scores, pioneering a promising direction for future explorations.

\bibliography{example_paper}
\bibliographystyle{icml2024}




\end{document}